\documentclass[10pt,twocolumn,letterpaper]{article}

\usepackage{cvpr}        
\usepackage[dvipsnames]{xcolor}

\definecolor{cvprblue}{rgb}{0.21,0.49,0.74}
\usepackage[pagebackref,breaklinks,colorlinks,citecolor=cvprblue]{hyperref}
\usepackage[accsupp]{axessibility}
\usepackage{booktabs}
\usepackage{multirow}
\usepackage{amsmath} 
\usepackage{caption} 
\usepackage{graphicx}

\definecolor{cvprblue}{rgb}{0.21,0.49,0.74}
\usepackage[pagebackref,breaklinks,colorlinks,citecolor=cvprblue]{hyperref}

\usepackage[capitalize]{cleveref}
\crefname{section}{Sec.}{Secs.}
\Crefname{section}{Section}{Sections}
\Crefname{table}{Table}{Tables}
\crefname{table}{Tab.}{Tabs.}

\newcommand{\pub}[1]{{\color{gray}{\tiny{[{#1}]}}}}

\def\confName{CVPR}
\def\confYear{2024}

\title{Event-assisted Low-Light Video Object Segmentation}

\author{
    Hebei Li \textsuperscript{1} \and
    Jin Wang \textsuperscript{1} \and
    Jiahui Yuan \textsuperscript{1} \and
    Yue Li \textsuperscript{1} \and
    Wenming Weng \textsuperscript{1} \and
    Yansong Peng \textsuperscript{1} \and
    Yueyi Zhang \textsuperscript{1,\thanks{Corresponding Author}} \and
    Zhiwei Xiong \textsuperscript{1} \and
    Xiaoyan Sun \textsuperscript{1, 2} \and
    \\
    \textsuperscript{1} University of Science and Technology of China \and
    \textsuperscript{2} Institute of Artificial Intelligence, Hefei Comprehensive National Science Center 
    \\
    \tt\small \{lihebei, jin01wang, yuanjiahui, yueli65, wmweng, pengyansong\}@mail.ustc.edu.cn,
    \\
    \tt\small \{zhyuey, zwxiong, sunxiaoyan\}@ustc.edu.cn
}

\begin{document}
\maketitle
\begin{abstract}
In the realm of video object segmentation (VOS), the challenge of operating under low-light conditions persists, resulting in notably degraded image quality and compromised accuracy when comparing query and memory frames for similarity computation. Event cameras, characterized by their high dynamic range and ability to capture motion information of objects, offer promise in enhancing object visibility and aiding VOS methods under such low-light conditions. This paper introduces a pioneering framework tailored for low-light VOS, leveraging event camera data to elevate segmentation accuracy. Our approach hinges on two pivotal components: the Adaptive Cross-Modal Fusion (ACMF) module, aimed at extracting pertinent features while fusing image and event modalities to mitigate noise interference, and the Event-Guided Memory Matching (EGMM) module, designed to rectify the issue of inaccurate matching prevalent in low-light settings. Additionally, we present the creation of a synthetic LLE-DAVIS dataset and the curation of a real-world LLE-VOS dataset, encompassing frames and events. Experimental evaluations corroborate the efficacy of our method across both datasets, affirming its effectiveness in low-light scenarios. The datasets are available at \url{https://github.com/HebeiFast/EventLowLightVOS}.
\end{abstract}    
\section{Introduction}
\label{sec:intro}

Video Object Segmentation (VOS) refers to a computer vision domain focusing on algorithms for segmenting objects within a sequence of video frames. Its applications span across various fields such as autonomous driving, surveillance, and interactive video editing \cite{zhang2016instance, wang2017selective}. VOS is commonly classified into two categories, contingent on the availability of annotation for the initial frame: semi-supervised and unsupervised VOS. Our research primarily delves into the realm of semi-supervised VOS.

Under low-light conditions, insufficient illumination gives rise to a diminished level of detail and compromised color accuracy. Conventional VOS methods \cite{cheng2021rethinking, yang2021associating, cheng2022xmem, li2022recurrent}, predicated upon the availability of high-quality visual inputs, tend to exhibit a marked decline in performance when confronted with such challenging lighting circumstances.

Unlike traditional imaging methods, event cameras signify a fundamental change and offer significant advantages in demanding lighting scenarios \cite{zhou2023deblurring, jiang2023event, liang2023coherent}. These cameras excel in high dynamic range and provide detailed edge and movement data for objects \cite{brandli2014240}. These unique features are crucial in segmenting video objects under low-light conditions. Despite their impressive capabilities, event cameras lack the ability of capture texture and color information, which limits their effectiveness in segmentation tasks. Hence, our paper aims to explore the integration of event-based and frame-based modalities to improve VOS, particularly in challenging low-light environments.

In pursuit of event-assisted low-light VOS, three pivotal challenges must be addressed: (i) The absence of a dedicated dataset tailored to low-light conditions for video object segmentation, encompassing both frames and events. Existing datasets are captured in standard lighting, leading to a significant gap in accurately representing low-light situations. (ii) Effectively exploiting complementary information from frame and event modalities under low-light conditions remains a complex task. Traditional approaches to integrate this data fall short due to inherent noise in low-light settings, necessitating the development of a more robust integration strategy for these modalities. (iii) Optimizing the utilization of event assistance for matching poses a substantial challenge. Current methods \cite{cheng2022xmem, yang2022decoupling, xu2022reliable} focus on enhancing the matching mechanism, assuming normal lighting conditions and relying solely on image data. Consequently, exploring effective means of leveraging event assistance in low-light scenarios becomes essential.

To tackle the above challenges, we build low-light datasets and introduce a novel end-to-end framework designed for low-light VOS, filling a critical research gap in VOS. For low-light VOS data, we construct a synthetic Low-Light Event DAVIS (LLE-DAVIS) dataset and collect a Low-Light Event Video Object Segmentation (LLE-VOS) dataset. These datasets serve as an important foundation for improving VOS techniques for low-light conditions. For our framework, we propose the adaptive cross-modal fusion (ACMF) module to combine complementary information by learning the adaptive filters to select useful information between two modalities. Besides, we propose an event-guided memory matching (EGMM) to solve the inaccurate matching mechanism. Our EGMM utilizes a joint approach of mask and event to guide the network in matching the target areas of memory, thereby enhancing the matching accuracy. We evaluate our method on the synthetic LLE-DAVIS dataset and LLE-VOS dataset. Experiments show our significant effectiveness, setting new standards for performance on both the LLE-DAVIS (62.8\%) and LLE-VOS (67.8\%) datasets.

In brief, our contributions are summarized as follows:
\begin{itemize}[leftmargin=*] \itemsep -1pt
\item We propose the first event-based low-light VOS framework by utilizing the unique properties of event cameras.
\item The first real-world event-based low-light VOS dataset is constructed, which contains event streams and frames captured under low-illumination scenarios, clear images and accurate annotations.
\item We elaborately design two components, i.e., an adaptive cross-modal fusion module and an event-guided memory matching module, for adaptively fusing the information of both frame and event modalities and enhancing the motion features for the matching module.
\item Both quantitative and qualitative results over synthetic and real-world datasets showcase that our proposed method outperforms existing state-of-the-art methods.
\end{itemize}
\section{Related Work}
\label{sec:related work}

\begin{figure*}
    \centering
    \includegraphics[width=1\linewidth]{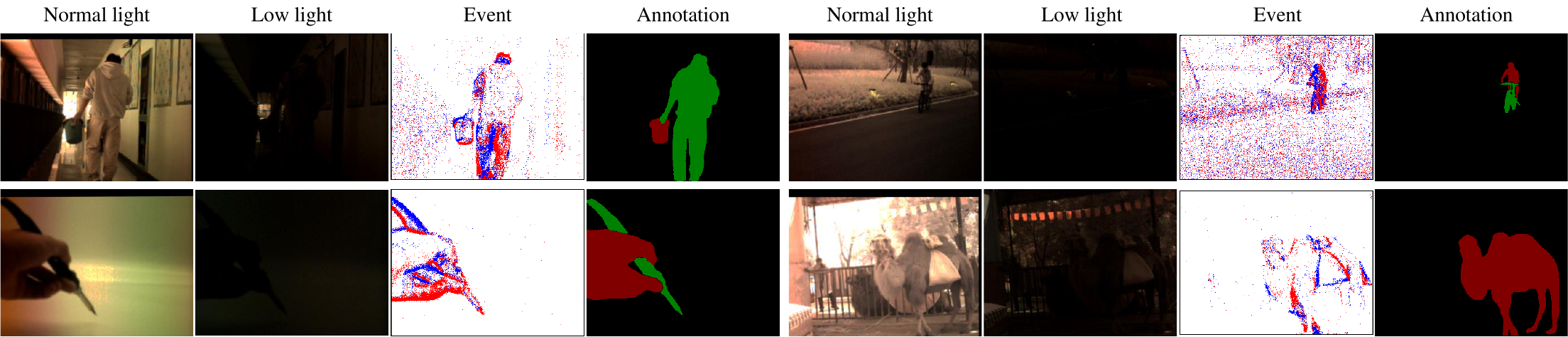}
    \caption{Examples of our LLE-VOS dataset. The dataset contains paired normal/low-light APS images, event stream and annotations.}
    \label{fig:example_dataset}
\end{figure*}
\begin{figure}
    \centering
    \includegraphics[width=1\linewidth]{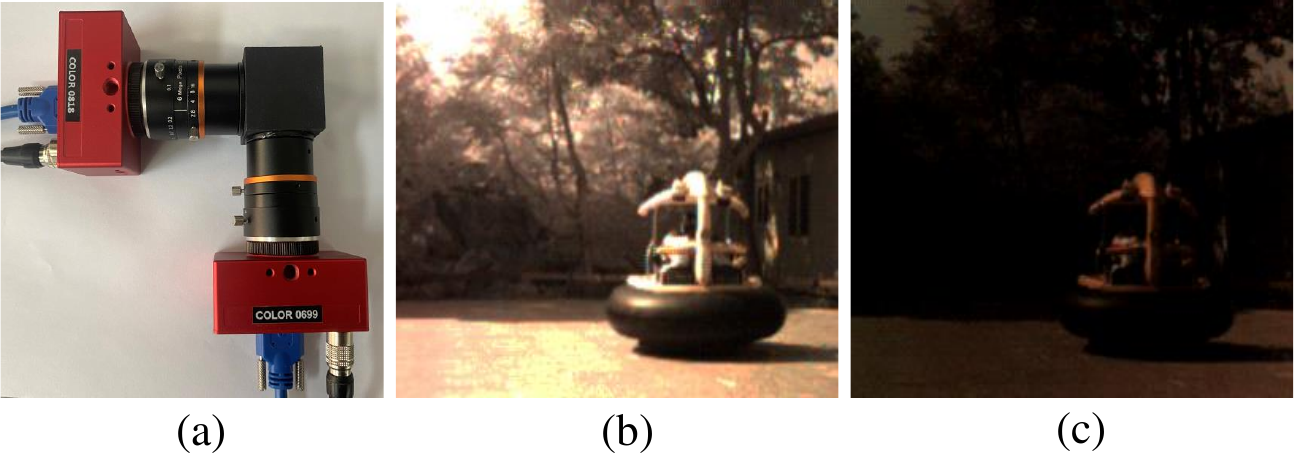}
    \caption{
    (a) A hybrid camera system for building real-world dataset.
    We configure two identical cameras with different exposure time for generating normal-light (b) and low-light (c) pairs.}
    \label{fig:device}
    \vspace{-10pt}
\end{figure}

\begin{table*}
    \centering
    \small
    \begin{tabular}{l|ccc|ccccc}
        \toprule
         & \multicolumn{3}{c|}{Synthetic LLE-DAVIS Dataset} & \multicolumn{5}{c}{Real-world LLE-VOS Dataset} \\
         \cline{2-9}
                          & Train & Validation & Total & Train & Validation & Val-Indoor & Val-Outdoor & Total \\
        \hline
\# Seq.  & 60 & 30 & 90 & 50 & 20 & 10 & 10 & 70 \\
\# Frm.  & 4149 & 1969 & 6118 & 3777 & 1823 & 993 & 830 & 5600 \\
\# Evt.  & 1406.19M & 626.70M & 2032.89M & 575.61M & 202.93M &                                  70.83M & 132.10M & 778.54M \\
Mean \# Frm. per Seq. & 69 & 66 & 68 & 76 & 91 & 99 & 83 & 80 \\

Mean \# Evt. per Seq. & 23.44M & 20.89M & 22.59M & 11.51M & 10.15M & 7.08M &                                 13.21M & 11.12M \\
Mean \# Objs. per Seq. & 2.4 & 2.0 & 2.3  & 1.8 & 1.6 & 1.7 & 1.5 & 1.7 \\
        \bottomrule
    \end{tabular}
    \caption{
    The summary of our synthetic LLE-DAVIS dataset and real-world LLE-VOS dataset, including the number of sequences (\#Seq.), frames (\#Frm.), events (\#Evt.), the mean number of frames, events and objects (\#Objs.) per sequence.}
    \label{tab:dataset}
    \vspace{-10pt}
\end{table*}
\subsection{Video Object Segmentation}
In the field of VOS, there are three categories of methods: online fine-tuning-based methods, propagation-based methods, and matching-based methods. Online fine-tuning-based methods \cite{caelles2017one, xiao2018monet, perazzi2017learning} concentrate on adjusting pre-existing segmentation networks during the evaluation phase to align them with the specific object being targeted. Propagation-based methods \cite{jampani2017video, chen2020state, xu2022reliable, oh2018fast, zhang2019fast} aim to expedite testing time by employing the mask from the previous frame to predict the mask for the current frame. Meanwhile, Matching-based methods, highlighted in \cite{oh2019video, seong2020kernelized, cheng2021rethinking, cheng2022xmem, yang2021associating, yang2022decoupling, mao2021joint, seong2021hierarchical}, ascertain pixel classification by evaluating its resemblance to the target object across memory frames.

\subsection{Event Segmentation}
There has been a growing interest in tailoring segmentation methods for event cameras, considering their advantageous features. Specifically, in event-based motion segmentation and event-based semantic segmentation. Stoffregen \emph{et al.} \cite{stoffregen2019event} proposed a distinctive per-event segmentation method aimed at estimating event-object associations. Mitrokhin \emph{et al.} \cite{mitrokhin2020learning} introduced a graph convolutional neural network to address the challenge of accurately analyzing dynamic scene evolution over time. Zhou \emph{et al.} \cite{zhou2021event} developed an approach to tackle motion segmentation in event-based camera data by minimizing energy and fitting multiple motion models. On the other hand, Alonso \emph{et al.} \cite{alonso2019ev} presented a novel representation of event data and released a new semantic dataset for event-based semantic segmentation. Additionally, Sun \emph{et al.} \cite{sun2022ess} proposed an unsupervised domain adaptation method for transferring semantic segmentation tasks from labeled image datasets to unlabeled event data. Moreover, Xia \emph{et al.} \cite{xia2023cmda} introduced an unsupervised cross-domain framework for effective nighttime semantic segmentation. Despite these advancements, a prevailing challenge lies in the inability of these methods to consistently track specific objects throughout entire video sequences. This limitation signifies a significant area necessitating further enhancement and future research endeavors.

\subsection{Low-Light Event Application}
Numerous studies \cite{shi2023even, zhou2023deblurring, liang2023coherent, liu2023low, zhou2021delieve} have delved into the potential of event cameras under low-light conditions. Zhang \emph{et al.} \cite{zhang2020learning} proposed an unsupervised domain adaptation network aimed at reconstructing images captured by event cameras in low-light conditions to resemble those taken during daylight. Jiang \emph{et al.} \cite{10168206} introduced a framework utilizing event cameras' superior dynamic range to produce clear images in near-darkness by integrating underexposed frames and event streams. Liu \emph{et al.} \cite{liu2023low} proposed a novel method for enhancing low-light videos using synthetic events from multiple frames, addressing the artifacts in extreme low light or fast-motion scenarios. Liang \emph{et al.} \cite{liang2023coherent} suggested a video enhancement approach for low-light conditions, establishing spatiotemporal coherence from frame-based and event cameras through a neural network. Zhou \emph{et al.} \cite{zhou2023deblurring} introduced a two-stage approach to enhance the deblurring of low-light images, leveraging the high dynamic range and low latency of event cameras. Nonetheless, prevailing research on event-based low-light scenarios primarily focuses on foundational tasks, leaving the domain of VOS largely unexplored.

\section{Benchmark Dataset}
\label{sec:dataset}
\begin{figure*}
    \centering
    \includegraphics[width=1\linewidth]{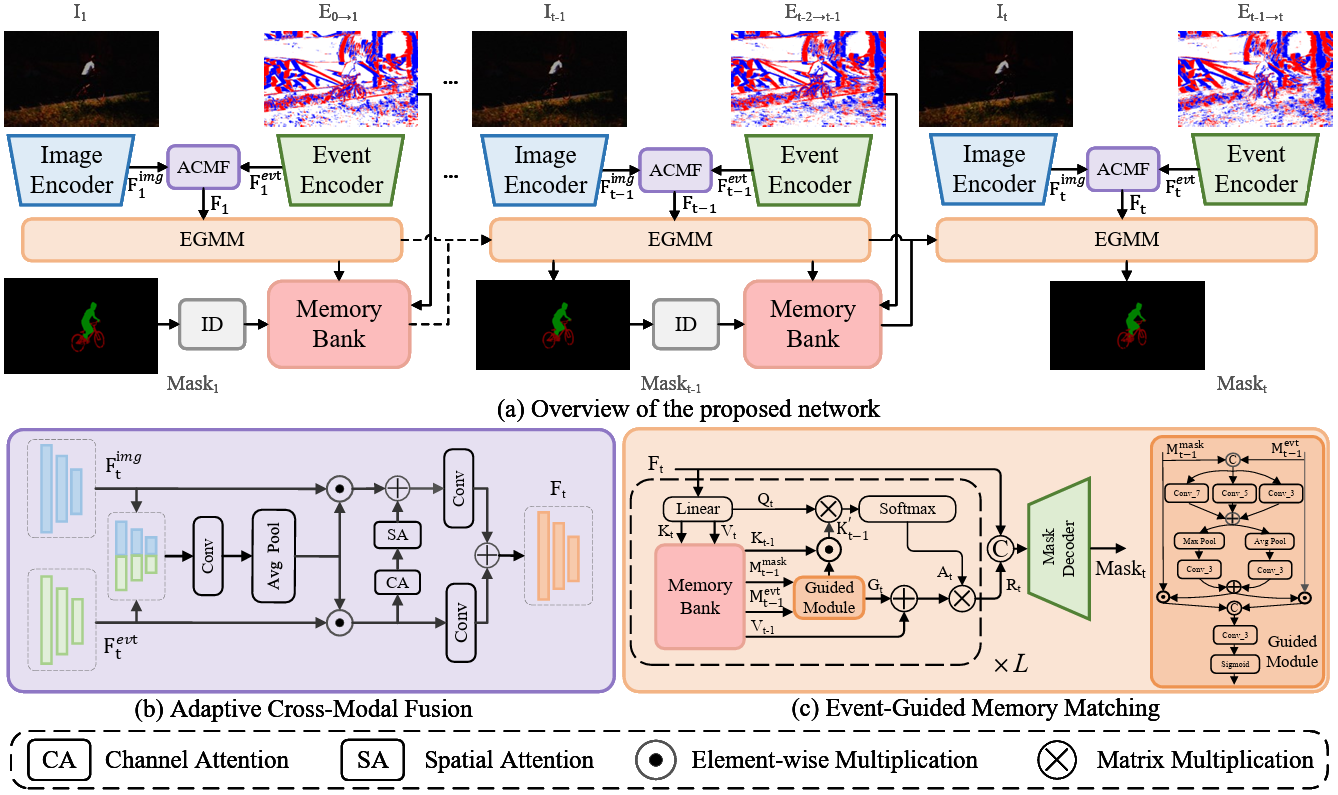}
    \caption{
    (a) Overview of the proposed method for event-assisted low-light video object segmentation.
 (b) The structure of Adaptive Cross-Modal Fusion (ACMF) module. (c) The structure of Event-Guided Memory Matching (EGMM) module.
    }
    \label{fig:main}
    \vspace{-8pt}
\end{figure*}

To the best of our knowledge, there has been no event-based VOS dataset in low-light scenarios. In this work, we build two low-light event-based VOS datasets, consisting of a synthetic LLE-DAVIS dataset and a real-world LLE-VOS dataset. \cref{tab:dataset} presents the summary of these two datasets, including the number of sequences, frames, objects, and events on each of the dataset splits. \cref{fig:example_dataset} shows some examples in the LLE-VOS dataset.

\subsection{Synthetic Dataset}
The DAVIS 2017 dataset is a widely used benchmark for the VOS task \cite{Oh_2019_ICCV, cheng2021rethinking, li2022recurrent, cheng2022xmem}. Thus based on this dataset, we construct a synthetic event dataset, named LLE-DAVIS, specifically tailored for the Low-Light Event-Based VOS task. Since the original DAVIS 2017 videos are recorded under normal-light conditions, we employ a devised technique \cite{lv2021attention} to generate low-light videos. Specifically, for a given normal-light frame $I_t$, we introduce random adjustments to the gamma correction factor and linear scaling factor to synthesize the corresponding low-light frame $L_t$. This process is mathematically expressed as:

\begin{equation}
\begin{aligned}
L_t = \beta \times (\alpha \times I_t)^{\gamma}+N_\sigma,
\end{aligned}
\end{equation}

Here, $\alpha$, $\beta$, $\gamma$ are randomly sampled from the uniform distributions $U(0.9, 1)$, $U(0.5, 1)$ and $U(7, 9)$, respectively. $N_{\sigma}$ represents Gaussian noise $N(0, \sigma)$, with $\sigma$ values drawn from a uniform distribution $U(0, 0.05)$. Subsequently, we employ FILM \cite{reda2022film} to interpolate DAVIS videos to 100fps. Then, these frames are fed into the ESIM model \cite{Gehrig_2020_CVPR} to generate events. In total, we have assembled 90 low-light video sequences, each accompanied by temporally-synchronized event streams.

\subsection{Real-World Dataset}
To collect the real-world LLE-VOS dataset, we build a hybrid camera system as shown in \cref{fig:device}(a).
The hybrid camera system is equipped with two DAVIS346 event cameras \cite{taverni2018front}, each adept at capturing temporally synchronized APS frames and event streams. 
A beam splitter is integrated into the system to uniformly distribute incoming light, guaranteeing that both cameras record identical scenes. 
Additionally, we conduct accurate geometric calibration and temporal synchronization between two cameras. 
More details are provided in the Supplementary Material. We adjust the exposure times of these two cameras to collect a pair of normal-light and low-light videos in one shot. We employ a group of 20 volunteers to accomplish the annotation work.

Our final dataset includes 70 videos, consisting of paired normal and low-light videos, along with their corresponding segmentation annotations and event streams. 
The videos are recorded at a diverse range of locations, including gyms, playgrounds, classrooms, meeting rooms, and zoos. To enhance the robustness of the dataset, it includes varying lighting conditions and contains a rich spectrum of motion information. We randomly select 50 video clips as the training set and 20 as the testing set. To further evaluate the robustness of our proposed method across various scenes, we divide the test set into indoor and outdoor scenarios, with 10 clips each. To our best, this is the first real-world dataset for low-light event-based VOS. It should be noted that this dataset can also be applied to other tasks such as event motion segmentation and low-light video enhancement.
\section{Method}                   \subsection{Overview}
We propose a novel event-assisted VOS framework, designed specifically for low-light conditions, as depicted in \cref{fig:main}(a). The core components of our architecture comprise an event encoder, an image encoder, an Event-Guided Memory Matching (EGMM) module, an Adaptive Cross-Modal Fusion (ACMF) module, and an Identity Assignment (ID) module \cite{yang2021associating}. The event encoder and image encoder extract the event feature and image feature, respectively. Then the multi-scale features of events and images are processed by the ACMF module, which generates complementary features. At the first time step, the EGMM module uses these features to produce query $Q_1$, key $K_1$ and value $V_1$. 
Subsequently, the ID module processes an initial segmentation annotation to create a mask feature. This generated mask feature, along with the predicted mask from the EGMM and the event features, are then systematically archived in the Memory Bank.
From the second time step until the final moment, the Memory Bank provides stored representations to the EGMM, along with the representations from image and event encoder. The EGMM then continually generates the mask at each time step, which is preserved in the Memory Bank for subsequent iterative refinement. 

\subsection{Event Representation}
Our proposed framework transforms an asynchronous event streams from $t-1$ to $t$ into corresponding voxel grids, denoted by $E_{t-1\rightarrow t}\in \mathbb{R}^{B\times H\times W}$, where $B$, $H$ and $W$ represent the number of temporal bins, the height and width of the grids, respectively. 

\subsection{Adaptive Cross-Modal Fusion}
\label{subsec: Adaptive Cross-Modal Fusion}
We introduce the Adaptive Cross-Modal Fusion (ACMF) to adaptively select the effect information from event and image in low-light environments, which is depicted in \cref{fig:main}(b). Under low-light conditions, the image modality provides inadequate texture and color information of objects while the event modality offers richer edge and motion cues. These distinct attributes of events are advantageous for segmentation tasks. 

Initially, ACMF merges the image feature $F_t^{img}$ and event feature $F_t^{evt}$ to create a more comprehensive feature set. This combined feature set is then processed to capture coupled information. The processed features act as a filter that selectively integrates information from two features through element-wise multiplications, reducing noise and enhancing relevant details. Diminished image quality under low-light conditions often results in lost structural details. Therefore, a channel-attention (CA) and a spatial-attention (SA) extract edge information from event feature to complete such details. This information helps to restore invisible object contours. Two convolutional layers further enhance two features. The final output is yielded by summing two features, acting as a robust representation for VOS.

\subsection{Event-Guided Memory Matching}
\label{subsec: Event-Guided Memory Matching}
Our proposed Event-Guided Memory Matching (EGMM) introduces a novel mechanism designed to enhance the accuracy of object matching between the current feature $F_t$ and the prior feature $K_{t-1}$, specifically under circumstances of imprecise mask predictions. The EGMM architecture, illustrated in \cref{fig:main}(b), integrates event feature with mask feature from memory to concentrate on motion-specific areas within a scene, consequently refining feature matching. 

Initially, the current feature $F_t$ generates the $Q_t$, $K_t$, $V_t$ through linear layer. $K_t$ and $V_t$ are saved in the Memory Bank for the next time step. Then, the event feature $M_{t-1}^{\text{evt}}$ and mask feature $M_{t-1}^{\text{mask}}$ from Memory Bank are sent to the Guided Module, showing in the right of \cref{fig:main}(c). The purpose of the guided branch is to improve unreliable mask predictions. It first integrates the multi-scale information from the concatenated feature through three parallel convolutional layers with various kernel sizes. Then, pooling layers and convolutional layers further reform its channel context, resulting in a guided signal for more accurate matching. Subsequently, the guided signal after passing a Sigmoid function is multiplicated with the past feature $K_{t-1}$, filtering misaligned features of the object region. The filtering process is described by the equation:
\begin{align}
    G_t &= \text{Guide}(M_{t-1}^{\text{evt}}, M_{t-1}^{\text{mask}}) \\
    K'_{t-1} &= K_{t-1} \cdot G_t,
\end{align}
where Guide$(.,.)$ represents the Guided module. $K_{t-1}, M_{t-1}^{\text{evt}}$ and $M_{t-1}^{\text{mask}}$ represent the previous key, event and mask features derived from memory, respectively. $K'_{t-1}$ and $G$ are the filtered key and guided branch output, respectively. 

The filtered key $K'_{t-1}$ interacts with $Q_t$ derived from $F_t$ through multiplication and softmax, formulated by:
\begin{equation}
A_t = \text{Softmax}\left(\frac{Q_t(K'_{t-1})^T}{\sqrt{d_k}}\right),
\end{equation}
where $A_t$ is the attention map and $d_k$ is the scaling factor corresponding to the dimension of the key vectors. 
The matching result $R_t$ is obtained by multiplication between the attention map $A_t$ and the summation of $G_t$ and $V_{t-1}$:
\begin{equation}
    R_t = A_t(G_t+V_{t-1}).
\end{equation}
Finally, we concat the current feature and matching result to combine the current and memory information. Then the concatenation result is sent to the mask decoder to generate the current mask.
\begin{equation}
Mask_{t} = Decoder(Concat(R_t, F_{t})).
\end{equation}

This EGMM module thereby provides a robust solution for enhancing the consistency of VOS, proving particularly beneficial in scenarios with unreliable mask predictions.

\begin{table*}
    \centering
    \begin{tabular}{cccccccccc}
    \toprule
    \multirow{2}{*}{Method}  & \multicolumn{3}{c}{Indoor Scenes} & \multicolumn{3}{c}{Outdoor Scenes} & \multicolumn{3}{c}{Overall} \\
        \cmidrule(lr){2-4} \cmidrule(lr){5-7} \cmidrule(lr){8-10}
    & $\mathcal{J}$ & $\mathcal{F}$ & $\mathcal{J}\&\mathcal{F}$ & $\mathcal{J}$ & $\mathcal{F}$ & $\mathcal{J}\&\mathcal{F}$ & $\mathcal{J}$ & $\mathcal{F}$ & $\mathcal{J}\&\mathcal{F}$ \\
    \hline
    STCN \pub{NIPS2021} \cite{cheng2021rethinking} & 0.486 & 0.321 & 0.403 & 0.400 & 0.309 & 0.354 & 0.445 & 0.316 & 0.380 \\
    XMem \pub{ECCV2022} \cite{cheng2022xmem} & 0.664 & 0.528 & 0.596 & 0.507 & 0.456 & 0.481 & 0.590 & 0.494 & 0.542 \\
    AOT \pub{NIPS2021} \cite{yang2021associating} & 0.699 & 0.618 & 0.659 & 0.592 & 0.571 & 0.581 & 0.649 & 0.596 & 0.623 \\
    DeAOT \pub{NIPS2022} \cite{yang2022decoupling} & 0.716 & 0.643 & 0.680 & 0.580 & 0.580 & 0.580 & 0.653 & 0.614 & 0.633 \\
    \hline
    Zero-DCE \cite{Zero-DCE} + STCN \cite{cheng2021rethinking}  & 0.513 & 0.334 & 0.424 & 0.415 & 0.313 & 0.364 & 0.467 & 0.324 & 0.396 \\
    Zero-DCE \cite{Zero-DCE} + XMem \cite{cheng2022xmem} & 0.681 & 0.527 & 0.604 & 0.535 & 0.497 & 0.516 & 0.612 & 0.513 & 0.563 \\
    Zero-DCE \cite{Zero-DCE} + AOT \cite{yang2021associating}  & 0.694 & 0.601 & 0.647 & 0.568 & 0.555 & 0.562 & 0.635 & 0.579 & 0.607 \\
    Zero-DCE \cite{Zero-DCE} + DeAOT \cite{yang2022decoupling} & 0.648 & 0.595 & 0.621 & 0.594 & 0.586 & 0.590 & 0.622 & 0.590 & 0.606 \\
    \hline
    Ours & \textbf{0.789} & \textbf{0.710} & \textbf{0.749} & \textbf{0.604} & \textbf{0.588} & \textbf{0.596} & \textbf{0.702} & \textbf{0.653} & \textbf{0.678}\\
    \bottomrule
        
    \end{tabular}
    \caption{Quantitative comparisons of various VOS methods on the real-world LLE-VOS dataset. The best results are marked in bold.}
    \label{tab:Real}
    \vspace{-8pt}
\end{table*}

\subsection{Loss function in VOS}
\label{subsec: Loss function in VOS}
To effectively train our VOS framework, we implement a composite loss function, denoted as $\mathcal{L}$. This function combines the Binary Cross-Entropy (BCE) loss and the Soft Jaccard (SJ) loss.  The overall loss is formalized as:
\begin{equation}
    \mathcal{L} = \sum_{t=1}^{T} \left( \alpha \sum_{o=1}^{N} \mathcal{L}_{\text{BCE}}^{(t,o)} + \beta \sum_{o=1}^{N} \mathcal{L}_{\text{SJ}}^{(t,o)} \right),
\end{equation}
where $\mathcal{L}_{\text{BCE}}^{(t,o)}$ and $\mathcal{L}_{\text{SJ}}^{(t,o)}$ represent the BCE loss and SJ loss for the object $o$ at time step $t$, respectively. $N$ denotes the total number of segmented objects. The parameters $\alpha$ and $\beta$ are the weights attributed to the BCE loss and IoU loss, respectively. $T$, $\alpha$, $\beta$ are set as 5, 0.5 and 0.5, respectively.
\section{Experiment}
\label{sec:Experiment}

\subsection{Comparison Methods}
We compare our method with state-of-the-art VOS methods, which include: STCN~\cite{cheng2021rethinking}, XMem~\cite{cheng2022xmem}, AOT~\cite{yang2021associating}, and DeAOT~\cite{yang2022decoupling}.
In addition to these direct VOS methods, we also compare our method with two-step approachs, which first utilize Zero-DCE~\cite{Zero-DCE} for low-light video enhancement and then apply above mentioned methods for VOS. 

Following~\cite{cheng2021rethinking, cheng2022xmem}, the evaluation of the VOS task employs the $\mathcal{J}$ score for segmentation accuracy, reflecting the average IoU between predictions and ground truth, and the $\mathcal{F}$ score for boundary preciseness, comparing the similarity of segmentation edges to actual contours. The mean of these scores denoted as $\mathcal{J}\&\mathcal{F}$, offers a comprehensive and balanced measure of overall performance.

\subsection{Implementation Details}
For model optimization, we deploy the AdamW optimizer with an initial learning rate of $2\times 10^{-4}$ and a weight decay of 0.07. We set our batch size to 8 and train our model over 50,000 iterations. For a fair comparison, we maintain the original training strategies of the models to obtain the best models. We apply standard data augmentation techniques: random scaling, random cropping, random horizontal flipping, and resizing. Besides, we randomly reverse the video and event sequences. The crop size of random cropping is (465, 465) for the LLE-DAVIS dataset and (256, 256) for the LLE-VOS dataset. All models are trained from scratch on both synthetic and real-world datasets under the same settings to ensure equitable comparisons. All the training experiments are conducted on four NVIDIA A800 GPUs.


\subsection{Experimental Results}
\subsubsection{Quantitative Results}

\textbf{Synthetic Dataset.}  \cref{tab:DAVIS} provides a detailed quantitative analysis of our framework in comparison with other methods on the LLE-DAVIS dataset. The comparative results clearly demonstrate that our end-to-end approach combined with event and image consistently outperforms the existing methods in terms of standard VOS metrics. Notably, our method exhibits a substantial performance increase with an improvement of 6.9\% over the AOT method in terms of $\mathcal{J}\&\mathcal{F}$. Additionally, our method significantly improves upon the two-stage method Zero-DCE+AOT, with increments of 6.7\% for $\mathcal{J}\&\mathcal{F}$. These improvements are due to the robust feature extraction capabilities of our framework and its effective integration of both image and event data in our VOS framework.

\begin{table}
    \centering
    \begin{tabular}{cccc}
    \toprule
    \multirow{2}{*}{Method} & \multicolumn{3}{c}{LLE-DAVIS}\\
    \cmidrule(lr){2-4}
    & $\mathcal{J}$ & $\mathcal{F}$ & $\mathcal{J}\&\mathcal{F}$  \\
    \hline
    STCN \pub{NIPS2021} \cite{cheng2021rethinking} & 0.424 & 0.453 & 0.438 \\
    XMem \pub{ECCV2022} \cite{cheng2022xmem} & 0.465 & 0.477 & 0.471 \\
    AOT  \pub{NIPS2021} \cite{yang2021associating} & 0.540 & 0.578 & 0.559 \\
    DeAOT \pub{NIPS2022} \cite{yang2022decoupling} & 0.541 & 0.571 & 0.556 \\
    \hline
    Zero-DCE \cite{Zero-DCE} + STCN \cite{cheng2021rethinking} & 0.440 & 0.469 & 0.455 \\
    Zero-DCE \cite{Zero-DCE} + XMem \cite{cheng2022xmem} & 0.494 & 0.512 & 0.503 \\
    Zero-DCE \cite{Zero-DCE} + AOT  \cite{yang2021associating}  & 0.544 & 0.577 & 0.561 \\
    Zero-DCE \cite{Zero-DCE} + DeAOT \cite{yang2022decoupling} & 0.541 & 0.577 & 0.559 \\
    \hline
    Ours & \textbf{0.602} & \textbf{0.654} & \textbf{0.628} \\
    \bottomrule
        
    \end{tabular}
    \caption{
    Quantitative comparisons of various VOS methods on the synthetic LLE-DAVIS dataset.
    The best results are marked in bold.
    }
    \label{tab:DAVIS}
    \vspace{-10pt}
\end{table}
\begin{figure*}[h!]
    \centering
    \includegraphics[width=1\textwidth]{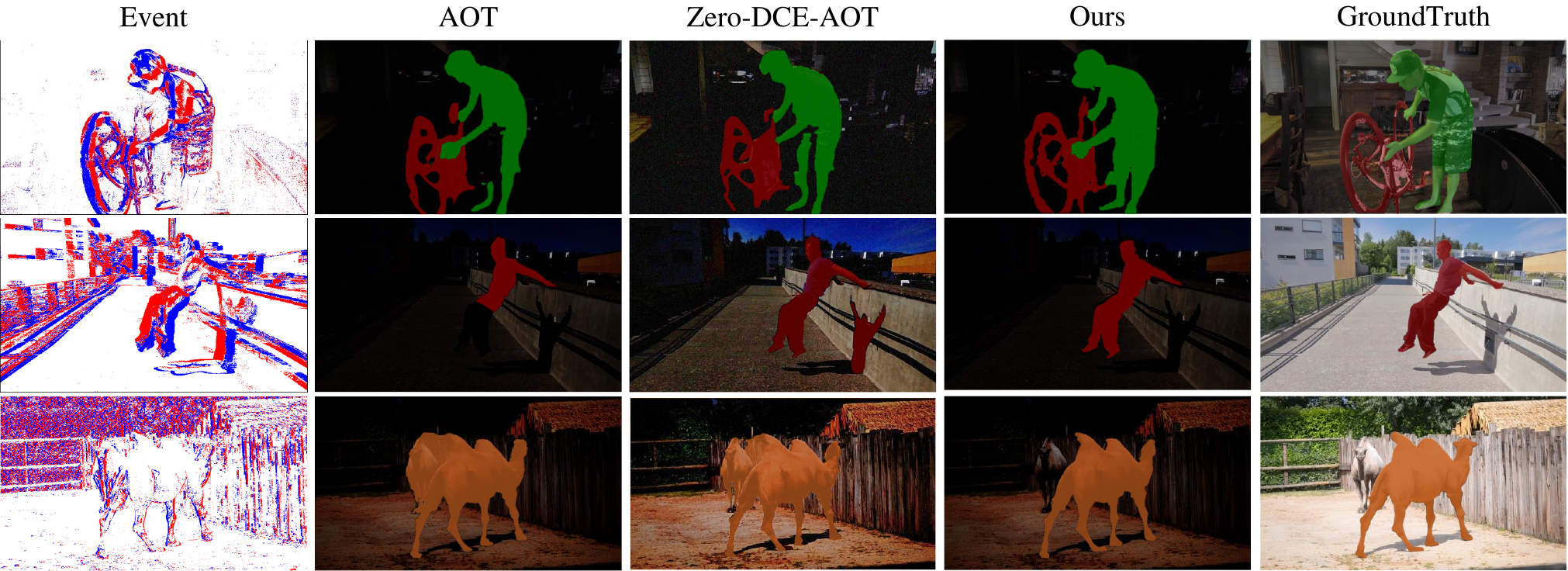}
    \caption{
    Qualitative comparisons with other methods on the synthetic LLE-DAVIS dataset.}
    \label{fig:Qual_DAVIS}
    \vspace{-5pt}
\end{figure*}

\begin{figure*}[h!]
    \centering
    \includegraphics[width=1\textwidth]{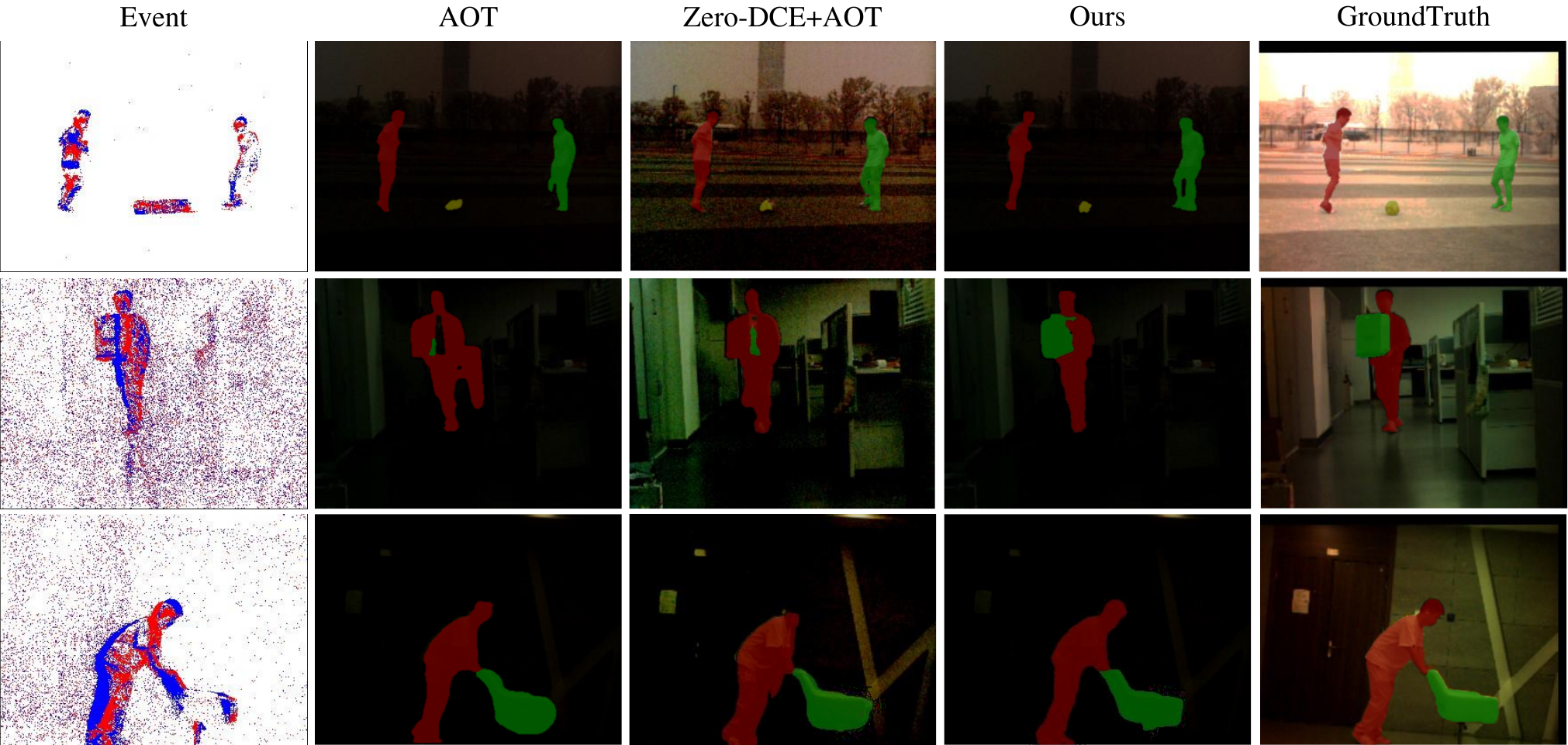}
    \caption{Qualitative comparisons with other methods on the real-world LLE-VOS dataset.}
    \label{fig:Qual_Real}
\end{figure*}

\noindent \textbf{Real-world Dataset.}
\cref{tab:Real} presents the quantitative results of one-stage and two-stage VOS methods on the LLE-VOS dataset. Given that real-world scenarios often involve extremely low-light conditions, the Zero-DCE method struggles to perform effectively. Therefore, the two-stage method Zero-DCE+AOT does not perform better than the AOT method. However, our method can improve the $\mathcal{J}\&\mathcal{F}$ metric over the existing one-stage VOS methods, with an increment when compared to DeAOT. Furthermore, our method outperforms Zero-DCE+AOT. In particular, it increases the $\mathcal{J}\&\mathcal{F}$ score from 0.607 to 0.678 in comparison to Zero-DCE+AOT. 
Our approach is measured against other one-stage and two-stage methods.

\subsection{Qualitative Results}
\cref{fig:Qual_DAVIS} showcases a qualitative comparison of our proposed method against both one-stage and two-stage methods on the synthetic LLE-DAVIS dataset under low-light conditions.  A key observation is that our method produces precise object masks that closely match the groundtruth. Instead, AOT either misses details or overlays redundant masks onto areas without objects. The Zero-DCE+AOT method offers enhancements over AOT alone, yet it fails to achieve the same level of clarity and precision provided by our method. These examples illustrate our method's superior ability to distinguish and outline objects in conditions where light is limited, demonstrating the practical advantage of integrating image and event data. This integration proves especially beneficial in difficult lighting, ensuring our VOS framework remains effective and reliable.

\noindent \textbf{Real-world Dataset.} \cref{fig:Qual_Real} illustrates qualitative results between our method and other approaches AOT and Zero-DCE+AOT on the real-world LLE-VOS dataset. The results of AOT method often miss parts of the objects, but our method is much better at finding the full shape, even when objects are hard to see in the frame. Our method stands out against Zero-DCE+AOT by being more precise and not mixing up different objects. This capability demonstrates that our approach is not only effective in simulated conditions but also maintains its reliability on real-world data, proving its practical applicability in low-light conditions. 

\begin{table}
    \centering 
    \begin{tabular}{c|c|ccc}
    \toprule
    & Method & $\mathcal{J}$ & $\mathcal{F}$ & $\mathcal{J}\&\mathcal{F}$ \\
    \midrule
    \multirow{3}{*}{\rotatebox[origin=c]{90}{\footnotesize Fwk Input}} 
    & Image Only & 0.540 & 0.578 & 0.559 \\
    & Event Only & 0.532 & 0.563 & 0.547 \\
    & Image + Event & \textbf{0.602} & \textbf{0.654} & \textbf{0.628} \\
    \midrule
    \multirow{3}{*}{\rotatebox[origin=c]{90}{\footnotesize Fwk Mod.}} 
    & Baseline & 0.555 & 0.614 & 0.584 \\
    & + ACMF & 0.578 & 0.623 & 0.601 \\
    & + ACMF + EGMM & \textbf{0.602} & \textbf{0.654} & \textbf{0.628} \\
    \midrule
    \multirow{4}{*}{\rotatebox[origin=c]{90}{\footnotesize EGMM Input}} 
    & None & 0.578 & 0.623 & 0.601 \\
    & Mask & 0.461 & 0.468 & 0.464 \\
    & Event & 0.540 & 0.563 & 0.552 \\
    & Event + Mask & \textbf{0.602} & \textbf{0.654} & \textbf{0.628} \\
    \bottomrule
    \end{tabular}
    \caption{Ablation results of different configurations on the VOS task. `Fwk Input', `Fwk Module' and `EGMM Input' represent the input type of framework input, framwork module and the input type of EGMM module, respectively. The best results are marked in bold.}
    \label{tab:Ablation}
    \vspace{-5pt}
\end{table}

\subsection{Ablation Study}
\textbf{Impact of input modalities.}
In the first part of \cref{tab:Ablation}, we show the comparative performance of the VOS network with different inputs. Utilizing only the image modality as input, the network achieves a $\mathcal{J}\&\mathcal{F}$ score of 0.559. In contrast, relying solely on the event modality yields a $\mathcal{J}\&\mathcal{F}$ score of 0.547. These results are inferior compared to the combined input model, which attains a $\mathcal{J}\&\mathcal{F}$ score of 0.628. This marked improvement through fusing both image and event modalities validates the necessity of multi-modality fusion for VOS under low-light conditions. 

\begin{table}
    \centering
    \begin{tabular}{c|cccccc}
    \toprule
        $L$ & $\mathcal{J}$ & $\mathcal{F}$ & $\mathcal{J}\&\mathcal{F}$ & FPS \\
        \hline
        1     & 0.577 & 0.628 & 0.603 & 22.50 \\
        2     & 0.595 & 0.642 & 0.618 & 21.32 \\
        3     & \textbf{0.602} & \textbf{0.654} & \textbf{0.628} & 20.26 \\
        4     & 0.594 & 0.649 & 0.622 & 19.16 \\
    \bottomrule
    \end{tabular}
    \caption{
    Ablation results of different numbers of EGMM blocks on the LLE-DAVIS dataset.
    The best results are marked in bold.}
    \label{tab:Ablation_EGMM}
\end{table}

\noindent \textbf{The effectiveness of ACMF and EGMM.} The second part of \cref{tab:Ablation} validates the effectiveness of the ACMF and EGMM modules within our framework. Initially, our baseline method employs a straightforward fusion approach by concatenating event and image features. The integration of the ACMF module enhances performance, improving the $\mathcal{J}$ score to 0.578, the $\mathcal{F}$ score to 0.623, and the composite $\mathcal{J}\&\mathcal{F}$ score to 0.601. Incorporating both ACMF and EGMM modules further boosts the 
$\mathcal{J}\&\mathcal{F}$ score to 0.628, indicating a significant contribution to the VOS performance on the synthetic LLE-DAVIS dataset.

\begin{figure}[t]
    \centering
    \includegraphics[width=1.0\linewidth]{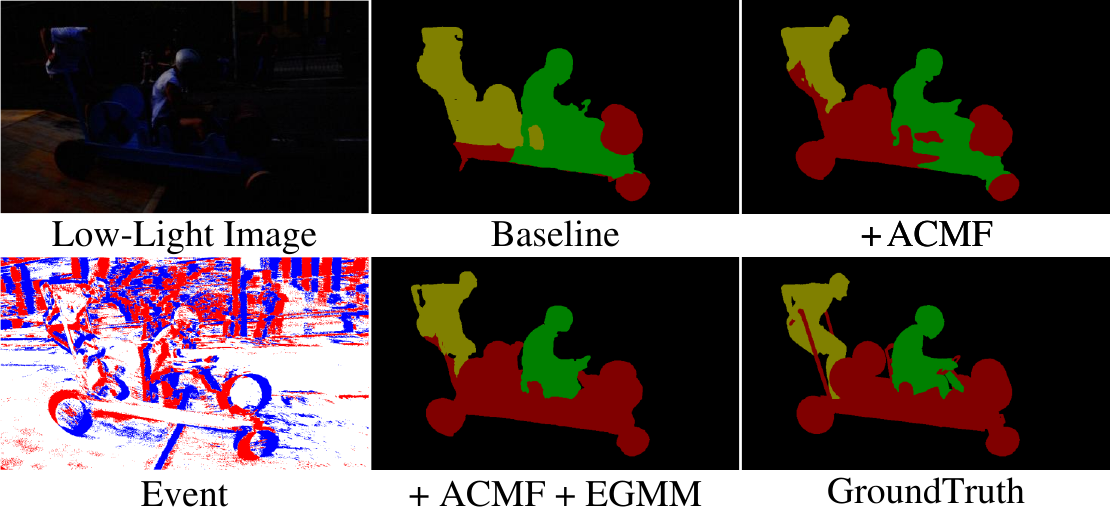}
    \caption{
    Visual results of ablation on different modules.}
    \label{fig:Ablation}
    \vspace{-12pt}
\end{figure}

\cref{fig:Ablation} presents a qualitative assessment of the individual and combined impacts of ACMF and EGMM. The baseline method exhibits inaccuracies in contour delineation and mask separation. The ACMF module refines the contour accuracy for the yellow human. However, the network still struggles to differentiate between the green human and the vehicle. The integration of the EGMM module effectively enables the network to distinguish between car and human.

\noindent
\textbf{The effect of event and mask in EGMM.}
To evaluate the individual and combined contributions of event and mask priors within the EGMM module, we present a quantitative analysis in the third part of \cref{tab:Ablation}. Without EGMM, the $\mathcal{J}\&\mathcal{F}$ score is 0.601.
Utilizing mask information alone yields a $\mathcal{J}\&\mathcal{F}$ score of 0.464 due to imprecise prediction to disturb matching. When the model processes only event data, the $\mathcal{J}\&\mathcal{F}$ is 0.552 due to noise and background motion. However, the fusion of both event and mask priors significantly enhances model performance, as evidenced by the improved $\mathcal{J}\&\mathcal{F}$ of 0.628. These results clearly illustrate the collaborate effect of combining mask and event information, which leads to a more robust and accurate performance in the EGMM.

\noindent
\textbf{The choice of EGMM block number.}
\cref{tab:Ablation_EGMM} examines the impact of changing the number of EGMM blocks on our model's performance. As we increase the number of EGMM blocks, the performance of our model improves, but it leads to more redundancy. When the number of EGMM blocks is four, the $\mathcal{J}\&\mathcal{F}$ score decreases. Therefore, we use three EGMM blocks in our experimental setup.

\section{Conclusion}
In this paper, we present a new approach to Video Object Segmentation (VOS) in low-light conditions. Unlike traditional VOS methods that rely heavily on high-quality video, our innovative framework uses the unique features of event cameras to improve segmentation accuracy. Our introduction of Adaptive Cross-Modal Fusion and Event-Guided Memory Matching has notably enhanced VOS performance. Our thorough testing on specially created Low-Light Event DAVIS (LLE-DAVIS) and Low-Light Event Video Object Segmentation (LLE-VOS) datasets proves that our method is superior, setting new standards for low-light VOS. This study paves the way for further breakthroughs in VOS and related fields.
\section{Acknowledgments}
This work was in part supported by the National Natural Science Foundation of China (NSFC) under grants 62032006 and 62021001.
{
    \small
    \bibliographystyle{ieeenat_fullname}
    \bibliography{main}

\begin{thebibliography}{40}
\providecommand{\natexlab}[1]{#1}
\providecommand{\url}[1]{\texttt{#1}}
\expandafter\ifx\csname urlstyle\endcsname\relax
  \providecommand{\doi}[1]{doi: #1}\else
  \providecommand{\doi}{doi: \begingroup \urlstyle{rm}\Url}\fi

\bibitem[Alonso and Murillo(2019)]{alonso2019ev}
Inigo Alonso and Ana~C Murillo.
\newblock Ev-segnet: Semantic segmentation for event-based cameras.
\newblock In \emph{Proceedings of the IEEE/CVF Conference on Computer Vision and Pattern Recognition Workshops}, 2019.

\bibitem[Brandli et~al.(2014)Brandli, Berner, Yang, Liu, and Delbruck]{brandli2014240}
Christian Brandli, Raphael Berner, Minhao Yang, Shih-Chii Liu, and Tobi Delbruck.
\newblock A 240 $\times$ 180 130 db 3 $\mu$s latency global shutter spatiotemporal vision sensor.
\newblock \emph{IEEE Journal of Solid-State Circuits}, 49\penalty0 (10):\penalty0 2333--2341, 2014.

\bibitem[Caelles et~al.(2017)Caelles, Maninis, Pont-Tuset, Leal-Taix{\'e}, Cremers, and Van~Gool]{caelles2017one}
Sergi Caelles, Kevis-Kokitsi Maninis, Jordi Pont-Tuset, Laura Leal-Taix{\'e}, Daniel Cremers, and Luc Van~Gool.
\newblock One-shot video object segmentation.
\newblock In \emph{Proceedings of the IEEE Conference on Computer Vision and Pattern Recognition (CVPR)}, pages 221--230, 2017.

\bibitem[Chen et~al.(2020)Chen, Li, Yuan, Yu, Shen, and Qi]{chen2020state}
Xi Chen, Zuoxin Li, Ye Yuan, Gang Yu, Jianxin Shen, and Donglian Qi.
\newblock State-aware tracker for real-time video object segmentation.
\newblock In \emph{Proceedings of the IEEE/CVF Conference on Computer Vision and Pattern Recognition (CVPR)}, pages 9384--9393, 2020.

\bibitem[Cheng and Schwing(2022)]{cheng2022xmem}
Ho~Kei Cheng and Alexander~G Schwing.
\newblock Xmem: Long-term video object segmentation with an atkinson-shiffrin memory model.
\newblock In \emph{European Conference on Computer Vision (ECCV)}, pages 640--658. Springer, 2022.

\bibitem[Cheng et~al.(2021)Cheng, Tai, and Tang]{cheng2021rethinking}
Ho~Kei Cheng, Yu-Wing Tai, and Chi-Keung Tang.
\newblock Rethinking space-time networks with improved memory coverage for efficient video object segmentation.
\newblock \emph{Advances in Neural Information Processing Systems}, 34:\penalty0 11781--11794, 2021.

\bibitem[Gehrig et~al.(2020)Gehrig, Gehrig, Hidalgo-Carri\'o, and Scaramuzza]{Gehrig_2020_CVPR}
Daniel Gehrig, Mathias Gehrig, Javier Hidalgo-Carri\'o, and Davide Scaramuzza.
\newblock Video to events: Recycling video datasets for event cameras.
\newblock In \emph{{IEEE} Conf. Comput. Vis. Pattern Recog. (CVPR)}, 2020.

\bibitem[Guo et~al.(2020)Guo, Li, Guo, Loy, Hou, Kwong, and Cong]{Zero-DCE}
Chunle~Guo Guo, Chongyi Li, Jichang Guo, Chen~Change Loy, Junhui Hou, Sam Kwong, and Runmin Cong.
\newblock Zero-reference deep curve estimation for low-light image enhancement.
\newblock In \emph{Proceedings of the IEEE conference on computer vision and pattern recognition (CVPR)}, pages 1780--1789, 2020.

\bibitem[Jampani et~al.(2017)Jampani, Gadde, and Gehler]{jampani2017video}
Varun Jampani, Raghudeep Gadde, and Peter~V Gehler.
\newblock Video propagation networks.
\newblock In \emph{Proceedings of the IEEE Conference on Computer Vision and Pattern Recognition (CVPR)}, pages 451--461, 2017.

\bibitem[Jiang et~al.(2023{\natexlab{a}})Jiang, Wang, Li, Zhang, Zhao, and Gao]{10168206}
Yu Jiang, Yuehang Wang, Siqi Li, Yongji Zhang, Minghao Zhao, and Yue Gao.
\newblock Event-based low-illumination image enhancement.
\newblock \emph{IEEE Transactions on Multimedia}, pages 1--12, 2023{\natexlab{a}}.

\bibitem[Jiang et~al.(2023{\natexlab{b}})Jiang, Wang, Li, Zhang, Zhao, and Gao]{jiang2023event}
Yu Jiang, Yuehang Wang, Siqi Li, Yongji Zhang, Minghao Zhao, and Yue Gao.
\newblock Event-based low-illumination image enhancement.
\newblock \emph{IEEE Transactions on Multimedia}, 26:\penalty0 1920--1931, 2023{\natexlab{b}}.

\bibitem[Li et~al.(2022)Li, Hu, Xiong, Zhang, Pan, and Liu]{li2022recurrent}
Mingxing Li, Li Hu, Zhiwei Xiong, Bang Zhang, Pan Pan, and Dong Liu.
\newblock Recurrent dynamic embedding for video object segmentation.
\newblock In \emph{Proceedings of the IEEE/CVF Conference on Computer Vision and Pattern Recognition (CVPR)}, pages 1332--1341, 2022.

\bibitem[Liang et~al.(2023)Liang, Yang, Li, Duan, Xu, and Shi]{liang2023coherent}
Jinxiu Liang, Yixin Yang, Boyu Li, Peiqi Duan, Yong Xu, and Boxin Shi.
\newblock Coherent event guided low-light video enhancement.
\newblock In \emph{Proceedings of the IEEE/CVF International Conference on Computer Vision (ICCV)}, pages 10615--10625, 2023.

\bibitem[Liu et~al.(2023)Liu, An, Liu, Yuan, Chen, Zhou, Li, Wang, and Tian]{liu2023low}
Lin Liu, Junfeng An, Jianzhuang Liu, Shanxin Yuan, Xiangyu Chen, Wengang Zhou, Houqiang Li, Yan~Feng Wang, and Qi Tian.
\newblock Low-light video enhancement with synthetic event guidance.
\newblock In \emph{Proceedings of the AAAI Conference on Artificial Intelligence}, pages 1692--1700, 2023.

\bibitem[Lv et~al.(2021)Lv, Li, and Lu]{lv2021attention}
Feifan Lv, Yu Li, and Feng Lu.
\newblock Attention guided low-light image enhancement with a large scale low-light simulation dataset.
\newblock \emph{International Journal of Computer Vision}, 129\penalty0 (7):\penalty0 2175--2193, 2021.

\bibitem[Mao et~al.(2021)Mao, Wang, Zhou, and Li]{mao2021joint}
Yunyao Mao, Ning Wang, Wengang Zhou, and Houqiang Li.
\newblock Joint inductive and transductive learning for video object segmentation.
\newblock In \emph{Proceedings of the IEEE/CVF International Conference on Computer Vision (ICCV)}, pages 9670--9679, 2021.

\bibitem[Mitrokhin et~al.(2020)Mitrokhin, Hua, Fermuller, and Aloimonos]{mitrokhin2020learning}
Anton Mitrokhin, Zhiyuan Hua, Cornelia Fermuller, and Yiannis Aloimonos.
\newblock Learning visual motion segmentation using event surfaces.
\newblock In \emph{Proceedings of the IEEE/CVF Conference on Computer Vision and Pattern Recognition (CVPR)}, pages 14414--14423, 2020.

\bibitem[Oh et~al.(2018)Oh, Lee, Sunkavalli, and Kim]{oh2018fast}
Seoung~Wug Oh, Joon-Young Lee, Kalyan Sunkavalli, and Seon~Joo Kim.
\newblock Fast video object segmentation by reference-guided mask propagation.
\newblock In \emph{Proceedings of the IEEE Conference on Computer Vision and Pattern Recognition (CVPR)}, pages 7376--7385, 2018.

\bibitem[Oh et~al.(2019{\natexlab{a}})Oh, Lee, Xu, and Kim]{Oh_2019_ICCV}
Seoung~Wug Oh, Joon-Young Lee, Ning Xu, and Seon~Joo Kim.
\newblock Video object segmentation using space-time memory networks.
\newblock In \emph{Proceedings of the IEEE/CVF International Conference on Computer Vision (ICCV)}, 2019{\natexlab{a}}.

\bibitem[Oh et~al.(2019{\natexlab{b}})Oh, Lee, Xu, and Kim]{oh2019video}
Seoung~Wug Oh, Joon-Young Lee, Ning Xu, and Seon~Joo Kim.
\newblock Video object segmentation using space-time memory networks.
\newblock In \emph{Proceedings of the IEEE/CVF International Conference on Computer Vision (ICCV)}, pages 9226--9235, 2019{\natexlab{b}}.

\bibitem[Perazzi et~al.(2017)Perazzi, Khoreva, Benenson, Schiele, and Sorkine-Hornung]{perazzi2017learning}
Federico Perazzi, Anna Khoreva, Rodrigo Benenson, Bernt Schiele, and Alexander Sorkine-Hornung.
\newblock Learning video object segmentation from static images.
\newblock In \emph{Proceedings of the IEEE Conference on Computer Vision and Pattern Recognition (CVPR)}, pages 2663--2672, 2017.

\bibitem[Reda et~al.(2022)Reda, Kontkanen, Tabellion, Sun, Pantofaru, and Curless]{reda2022film}
Fitsum Reda, Janne Kontkanen, Eric Tabellion, Deqing Sun, Caroline Pantofaru, and Brian Curless.
\newblock Film: Frame interpolation for large motion.
\newblock In \emph{European Conference on Computer Vision}, pages 250--266. Springer, 2022.

\bibitem[Seong et~al.(2020)Seong, Hyun, and Kim]{seong2020kernelized}
Hongje Seong, Junhyuk Hyun, and Euntai Kim.
\newblock Kernelized memory network for video object segmentation.
\newblock In \emph{Computer Vision--ECCV 2020: 16th European Conference, Glasgow, UK, August 23--28, 2020, Proceedings, Part XXII 16}, pages 629--645. Springer, 2020.

\bibitem[Seong et~al.(2021)Seong, Oh, Lee, Lee, Lee, and Kim]{seong2021hierarchical}
Hongje Seong, Seoung~Wug Oh, Joon-Young Lee, Seongwon Lee, Suhyeon Lee, and Euntai Kim.
\newblock Hierarchical memory matching network for video object segmentation.
\newblock In \emph{Proceedings of the IEEE/CVF International Conference on Computer Vision (ICCV)}, pages 12889--12898, 2021.

\bibitem[Shi et~al.(2023)Shi, Peng, Qiu, Ju, Lo, and Lo]{shi2023even}
Peilun Shi, Jiachuan Peng, Jianing Qiu, Xinwei Ju, Frank Po~Wen Lo, and Benny Lo.
\newblock Even: An event-based framework for monocular depth estimation at adverse night conditions.
\newblock \emph{arXiv preprint arXiv:2302.03860}, 2023.

\bibitem[Stoffregen et~al.(2019)Stoffregen, Gallego, Drummond, Kleeman, and Scaramuzza]{stoffregen2019event}
Timo Stoffregen, Guillermo Gallego, Tom Drummond, Lindsay Kleeman, and Davide Scaramuzza.
\newblock Event-based motion segmentation by motion compensation.
\newblock In \emph{Proceedings of the IEEE/CVF International Conference on Computer Vision (ICCV)}, pages 7244--7253, 2019.

\bibitem[Sun et~al.(2022)Sun, Messikommer, Gehrig, and Scaramuzza]{sun2022ess}
Zhaoning Sun, Nico Messikommer, Daniel Gehrig, and Davide Scaramuzza.
\newblock Ess: Learning event-based semantic segmentation from still images.
\newblock In \emph{European Conference on Computer Vision (ECCV)}, pages 341--357. Springer, 2022.

\bibitem[Taverni et~al.(2018)Taverni, Moeys, Li, Cavaco, Motsnyi, Bello, and Delbruck]{taverni2018front}
Gemma Taverni, Diederik~Paul Moeys, Chenghan Li, Celso Cavaco, Vasyl Motsnyi, David San~Segundo Bello, and Tobi Delbruck.
\newblock Front and back illuminated dynamic and active pixel vision sensors comparison.
\newblock \emph{IEEE Transactions on Circuits and Systems II: Express Briefs}, 65\penalty0 (5):\penalty0 677--681, 2018.

\bibitem[Wang et~al.(2017)Wang, Shen, and Porikli]{wang2017selective}
Wenguan Wang, Jianbing Shen, and Fatih Porikli.
\newblock Selective video object cutout.
\newblock \emph{IEEE Transactions on Image Processing}, 26\penalty0 (12):\penalty0 5645--5655, 2017.

\bibitem[Xia et~al.(2023)Xia, Zhao, Zheng, Wu, Sun, and Tang]{xia2023cmda}
Ruihao Xia, Chaoqiang Zhao, Meng Zheng, Ziyan Wu, Qiyu Sun, and Yang Tang.
\newblock Cmda: Cross-modality domain adaptation for nighttime semantic segmentation.
\newblock In \emph{Proceedings of the IEEE/CVF International Conference on Computer Vision (ICCV)}, pages 21572--21581, 2023.

\bibitem[Xiao et~al.(2018)Xiao, Feng, Lin, Liu, and Zhang]{xiao2018monet}
Huaxin Xiao, Jiashi Feng, Guosheng Lin, Yu Liu, and Maojun Zhang.
\newblock Monet: Deep motion exploitation for video object segmentation.
\newblock In \emph{Proceedings of the IEEE Conference on Computer Vision and Pattern Recognition (CVPR)}, pages 1140--1148, 2018.

\bibitem[Xu et~al.(2022)Xu, Wang, Li, and Lu]{xu2022reliable}
Xiaohao Xu, Jinglu Wang, Xiao Li, and Yan Lu.
\newblock Reliable propagation-correction modulation for video object segmentation.
\newblock In \emph{Proceedings of the AAAI Conference on Artificial Intelligence}, pages 2946--2954, 2022.

\bibitem[Yang and Yang(2022)]{yang2022decoupling}
Zongxin Yang and Yi Yang.
\newblock Decoupling features in hierarchical propagation for video object segmentation.
\newblock \emph{Advances in Neural Information Processing Systems}, 35:\penalty0 36324--36336, 2022.

\bibitem[Yang et~al.(2021)Yang, Wei, and Yang]{yang2021associating}
Zongxin Yang, Yunchao Wei, and Yi Yang.
\newblock Associating objects with transformers for video object segmentation.
\newblock \emph{Advances in Neural Information Processing Systems}, 34:\penalty0 2491--2502, 2021.

\bibitem[Zhang et~al.(2019)Zhang, Lin, Zhang, Lu, and He]{zhang2019fast}
Lu Zhang, Zhe Lin, Jianming Zhang, Huchuan Lu, and You He.
\newblock Fast video object segmentation via dynamic targeting network.
\newblock In \emph{Proceedings of the IEEE/CVF International Conference on Computer Vision (ICCV)}, pages 5582--5591, 2019.

\bibitem[Zhang et~al.(2020)Zhang, Zhang, Jiang, Zou, Ren, and Zhou]{zhang2020learning}
Song Zhang, Yu Zhang, Zhe Jiang, Dongqing Zou, Jimmy Ren, and Bin Zhou.
\newblock Learning to see in the dark with events.
\newblock In \emph{Computer Vision--ECCV 2020: 16th European Conference, Glasgow, UK, August 23--28, 2020, Proceedings, Part XVIII 16}, pages 666--682. Springer, 2020.

\bibitem[Zhang et~al.(2016)Zhang, Fidler, and Urtasun]{zhang2016instance}
Ziyu Zhang, Sanja Fidler, and Raquel Urtasun.
\newblock Instance-level segmentation for autonomous driving with deep densely connected mrfs.
\newblock In \emph{Proceedings of the IEEE Conference on Computer Vision and Pattern Recognition}, pages 669--677, 2016.

\bibitem[Zhou et~al.(2021{\natexlab{a}})Zhou, Teng, Han, Xu, and Shi]{zhou2021delieve}
Chu Zhou, Minggui Teng, Jin Han, Chao Xu, and Boxin Shi.
\newblock Delieve-net: Deblurring low-light images with light streaks and local events.
\newblock In \emph{Proceedings of the IEEE/CVF International Conference on Computer Vision}, pages 1155--1164, 2021{\natexlab{a}}.

\bibitem[Zhou et~al.(2023)Zhou, Teng, Han, Liang, Xu, Cao, and Shi]{zhou2023deblurring}
Chu Zhou, Minggui Teng, Jin Han, Jinxiu Liang, Chao Xu, Gang Cao, and Boxin Shi.
\newblock Deblurring low-light images with events.
\newblock \emph{International Journal of Computer Vision}, 131\penalty0 (5):\penalty0 1284--1298, 2023.

\bibitem[Zhou et~al.(2021{\natexlab{b}})Zhou, Gallego, Lu, Liu, and Shen]{zhou2021event}
Yi Zhou, Guillermo Gallego, Xiuyuan Lu, Siqi Liu, and Shaojie Shen.
\newblock Event-based motion segmentation with spatio-temporal graph cuts.
\newblock \emph{IEEE Transactions on Neural Networks and Learning Systems}, 34\penalty0 (8):\penalty0 4868--4880, 2021{\natexlab{b}}.

\end{thebibliography}
}


\end{document}